\pdfoutput=1
\documentclass[11pt,a4paper]{article}
\usepackage[T1]{fontenc}
\usepackage{lmodern}
\usepackage[margin=24mm]{geometry}
\usepackage{amsmath,amssymb,amsthm,bm}
\usepackage{graphicx,xcolor,booktabs,array,multirow}
\usepackage[font=small,labelfont=bf,labelsep=period]{caption}
\usepackage{microtype}
\usepackage{xspace}
\usepackage{siunitx}
\usepackage[numbers,square,sort&compress]{natbib}
\usepackage[colorlinks=true,linkcolor=black,filecolor=black,citecolor=red,urlcolor=blue!45!black]{hyperref}
\DeclareRobustCommand{\figref}[2]{\begingroup
\hypersetup{linkcolor=blue!70!black,filecolor=blue!70!black}%
\hyperref[#1]{\ref*{#1}#2}\endgroup}

\graphicspath{{figures/}}

\theoremstyle{definition}
\theoremstyle{remark}

\newcommand{\pp}{\ensuremath{\,\mathrm{pp}}\xspace}   %

\newcommand{\nOptObjTest}{1{,}797}
\newcommand{\nPenObjTest}{3{,}498}
\newcommand{\nMfeatObjTest}{1{,}000}

\newcommand{\nPenHardPairThreeBitGainPP}{+1.97}
\newcommand{\nMfeatHardPairThreeBitGainPP}{\ensuremath{-}0.61}

\newcommand{\nOptUnitSeedZeroLMPairJointAcc}{88.31}
\newcommand{\nOptUnitSeedZeroLMPairIndepAcc}{63.38}

\newcommand{\nOptGainNullMinusNatPP}{\ensuremath{-}0.55}
\newcommand{\nPenGainNullMinusNatPP}{+0.62}
\newcommand{\nMfeatGainNullMinusNatPP}{+0.41}

\newcommand{\nCtlRandDraws}{500}
\newcommand{\nCtlRandInjectiveDraws}{18}
\newcommand{\nCtlCrossAllIndependentTwoBitDistinctSymbolPairsMax}{9}
\newcommand{\nCtlZernMfeatHardTFourNFavoursZernike}{7}

\newcommand{\nCtlCrossOptSeedZeroLMIndepTwoBitNameable}{7}
\newcommand{\nCtlCrossAllJointTwoBitNUnitsNameAllTen}{23}
\newcommand{\nPoolHardPairTwoBitNPos}{24}
\newcommand{\nCtlConvOptTwoBitConventionFracJointGainRecoveredPct}{54.1}
\newcommand{\nCtlConvPenTwoBitConventionFracJointGainRecoveredPct}{43.6}
\newcommand{\nCtlConvMfeatTwoBitConventionFracJointGainRecoveredPct}{63.0}
\newcommand{\nOptGainTwoBitPPAbs}{22.68}
\newcommand{\nPenGainTwoBitPPAbs}{18.07}
\newcommand{\nMfeatGainTwoBitPPAbs}{22.41}

\newcommand{\nRevLibrarySize}{500}
\newcommand{\nRevPenCommonGain}{+2.51}
\newcommand{\nRevMfeatCommonGain}{+1.05}
\newcommand{\nRevBeanActivePredicted}{+1.46}
\newcommand{\nRevBeanActiveGain}{\ensuremath{-}0.24}
\newcommand{\nRevBeanActiveLower}{\ensuremath{-}0.71}
\newcommand{\nRevBeanActiveUpper}{+0.21}
\newcommand{\nLlmSeventyCoord}{23.08\xspace}
\newcommand{\nLlmSeventySoloGap}{-9.67\xspace}
\newcommand{\nLlmFineCoord}{30.49\xspace}
\newcommand{\nLlmHiddenFineLabelSolo}{87.56\xspace}
\newcommand{\nLlmFineQCount}{64\xspace}
\newcommand{\nLlmFineQPred}{1.69\xspace}
\newcommand{\nLlmFineQObserved}{-6.18\xspace}

\newcommand{\nNatSmallActivation}{5.86}
\newcommand{\nNatSmallReceiverAccuracy}{94.90}
\newcommand{\nNatLargeActivation}{90.49}
\newcommand{\nNatIdentityThreeRawAccuracy}{75.31}
\newcommand{\nRegPenTFourDiff}{+0.91}
\newcommand{\nRegLlmMaxGainAllLevels}{+0.58}
\newcommand{\nRegOptTSixErrRed}{28.85}
\newcommand{\nRegPenTSixErrRed}{20.46}
\newcommand{\nRegNatSmallCalibrationN}{32}
\newcommand{\nRegNatLargeCalibrationN}{512}
\newcommand{\nNsTopView}{96.51}
\newcommand{\nNsGainMaxAbs}{38.78}
\newcommand{\nSenClasses}{19}
\newcommand{\nSenPersons}{8}
\newcommand{\nSenUnits}{5}
\newcommand{\nSenGateAdmissible}{1}
\newcommand{\nSenDevBTwoNThreePTwoPoint}{+4.82}
\newcommand{\nSenTOneBTwoNTwoGainIndep}{+24.71}
\newcommand{\nSenTOneBTwoNThreeGainSolo}{+0.04}
\newcommand{\nSenPTwoTOnePrimaryCells}{6}
\newcommand{\nSenPTwoTOnePrimaryPredLossCells}{5}
\newcommand{\nSenPThreeARatioMin}{2.8}
\newcommand{\nSenPThreeARatioMax}{68.1}
\newcommand{\nSenPThreeACellsWithMass}{7}
\newcommand{\nSenPFourTOneArgmaxGainAbs}{23.99}
\newcommand{\nSenTeamTOneUnseenMassMax}{22.24}
\newcommand{\nEeSeeds}{3}
\newcommand{\nEeJointEncBTwoAcc}{95.18}
\newcommand{\nEeJointLabelBTwoAcc}{96.73}
\newcommand{\nEeJointEncBThreeAcc}{97.28}
\newcommand{\nEeWsJointEncBThreeAcc}{97.46}
\newcommand{\nEeDiffJointEncMinusJointLabelBTwoMeanAbs}{1.55}
\newcommand{\nEeDiffJointEncMinusSoloBThreeMean}{+2.67}
\newcommand{\nEeErrRedJointEncVsSoloBThree}{49.54}

\newcommand{\nCvGateThreshold}{80}
\newcommand{\nCvCubClasses}{200}
\newcommand{\nCvWosClasses}{134}
\newcommand{\nCvCubCaptionAcc}{41.43}
\newcommand{\nCvCubImageAcc}{85.12}
\newcommand{\nCvWosAbstractAcc}{60.71}
\newcommand{\nCvWosKeywordAcc}{52.19}

\newcommand{\nCiJeSoloSixLower}{2.32}
\newcommand{\nCiJeSoloSixUpper}{3.04}

\makeatletter
\newcommand{\topalignedfigurepages}{%
  \setlength{\@fptop}{0pt}%
  \setlength{\@fpbot}{0pt plus 1fil}}
\makeatother

\newcommand{\papertitle}{The Tower of Babel in artificial intelligence: who should speak, and when it pays to listen}
\newcommand{\paperauthor}{Mohammad Salahshour\\[4pt]
{\small\itshape Department of Collective Behaviour, Max Planck Institute of Animal Behavior, Konstanz, Germany}\\
{\small\itshape Centre for the Advanced Study of Collective Behaviour, University of Konstanz, Konstanz, Germany}\\
{\small\itshape Department of Biology, University of Konstanz, Konstanz, Germany}\\[2pt]
{\small\texttt{salahshour.mohammad@gmail.com}}}
\begin{document}
\topalignedfigurepages
\title{\papertitle}
\author{\paperauthor}
\date{}
\maketitle

\begin{abstract}
Two machines can each recognize an object well and still fail together. We study classifiers that see different parts of one object and send short messages to a receiver that identifies its class. Coordination asks how to design messages together; cooperation asks whether the team outperforms one sender given the whole budget. On three digit datasets, messages designed together beat messages designed apart by \nPenGainTwoBitPPAbs{} to \nOptGainTwoBitPPAbs{} percentage points on average with two-bit messages, with gains in all \nPoolHardPairTwoBitNPos{} configurations. A fixed agreement, you report the row and I the column, recovers about half this gain, with only the receiver learned. Senders that draw the same distinctions waste a narrow vocabulary, and senders whose distinctions cross multiply it. A team has no more possible messages than one sender with the same budget, so a partner must bring evidence the first lacks. A solvable model shows individual statistics alone cannot determine whether a partnership helps. On MNIST digits, six shared bits remove \nEeErrRedJointEncVsSoloBThree{} percent of the strongest tested solo's errors on average. Paired measurements correctly predicted most sensor teams would reduce average accuracy on new participants. Who should speak is a property of the pair, and must be measured in pairs.
\end{abstract}
\section{Introduction}\label{sec:intro}

In the old story, the tower of Babel is abandoned when its builders can no longer understand one another. Every collective that must pool what its members know meets a version of that difficulty. Animal groups combine observations made by different individuals, and human teams combine abilities that no member has in full~\citep{Couzin2009Collective,Woolley2010Collective}. What a group achieves depends on how its members' information is used, as well as on how much each member holds~\citep{Salahshour2019}. The same question arises in teams of artificial systems. Frameworks assemble teams of language-model agents~\citep{Wu2023}, and exchanges between agents improve answers on some reasoning tasks~\citep{Du2023}. Yet multi-agent systems fail in characteristic ways~\citep{Cemri2025}, a single agent with the same budget often matches or beats the team~\citep{Tran2026Single}, and human--machine teams often perform below their stronger member~\citep{Vaccaro2024}.

The members of such a collective hold private representations shaped by distinct views and histories~\citep{SalahshourPerceptual2025}. Even when members establish common ground on the meanings of task outcomes~\citep{Salahshour2026gauge,Salahshour2020}, a fundamental bottleneck remains. Cells within an organism and animals in a group coordinate through compressed signals~\citep{Ellison2016Communication,StrandburgPeshkin2013Visual}, just as sensors, neural modules, and AI agents must exchange information under strict limits on bandwidth, time, or token budgets~\citep{Tran2026Single,Goyal2022Workspace,Foerster2016,Gunduz2023,Lemasson2013Attention}. A narrow channel that carries only a few bits forces each member to compress what it knows into coarse categories, and forces the group to decide whether sharing the channel beats giving the entire budget to one capable speaker.

We treat it as a game of common interest. The members share one goal and hold private information and private codes, as in the signalling games of Lewis and the team problems of Marschak and Radner~\citep{Lewis1969,MarschakRadner1972,Skyrms2010}. Without any conflict of interest, two problems remain. One is coordination. Which convention should divide the vocabulary between the senders, so that a receiver can use their messages together? The other is cooperation in the sense of resources. Is a partner's share of the channel worth more than the same bits in the hands of one capable member? In this game, cooperation carries no dilemma: the senders share a goal, and the cost of a partner is the share of the channel it takes. Decentralized detection designs the reports of several sensors jointly and, under a shared rate, has asked how many identical sensors should speak and with how many bits each~\citep{Tenney1981,Tsitsiklis1993,Viswanathan1997,Chamberland2003,Longo1990}. Coding for computing preserves the information that one answer needs~\citep{Orlitsky2001,Doshi2010}, and task-oriented and emergent communication extend both to learned systems~\citep{Gunduz2023,Shao2022,Foerster2016,Lazaridou2020}. For systems learned from data, we must make this decision from samples of senders that may differ substantially. We ask how to determine, before deployment, whether sharing the budget improves on giving it to one capable sender.

To isolate the mechanics of bandwidth sharing from the confounders of free-text generation and prompt heuristics, we study communication through controlled discrete bottlenecks. Classifiers and supervised readouts of frozen language and vision models observe different views of the same object and send short bit-limited messages to a receiver that predicts the object's class. Holding the encoders' underlying representations fixed while varying only their message rules and bit allocations separates communication design from individual capability. Alongside an exactly solvable model of synergistic evidence, we evaluate teams across handwritten digits, hierarchical text abstracts, tabular morphology, multi-modal bird and research-paper archives, and wearable sensor recordings.

We find a sharp divergence between learning to speak together and beating a solo speaker. Coordination is large and cheap, with capacity limits set by combinatorial counting: at representative budgets, joint code design raises team accuracy by 18 to 39 percentage points across domains, and on the digit datasets a data-blind convention recovers roughly half of that improvement without learning the encoders. Yet large coordination gains routinely mask cooperation failures. In our solvable model, identical individual statistics can make partnership either useless or error-free depending on how observations couple. Empirically, sharing the bit budget beats the strongest tested equal-budget solo on complementary digit views---eliminating \nEeErrRedJointEncVsSoloBThree\% of solo errors on average at six total bits on split MNIST digits---while often losing to or tying the solo on text halves, tabular features, and wearable sensors despite massive coordination gains. Furthermore, predictions of team benefit derived from separate member statistics mistakenly approve partnerships on text and dry-bean tasks, whereas paired calibration on the same objects correctly predicts whether sensor teams improve or degrade average performance on new participants. Who should speak is an irreducible property of the pair, and we identify the comparisons and measurements needed to assess it.
\section{Results}\label{sec:results}

\subsection{A game of common interest with a narrow channel}\label{sec:m-game}

\begin{figure}[!p]
\centering
\includegraphics[width=\textwidth]{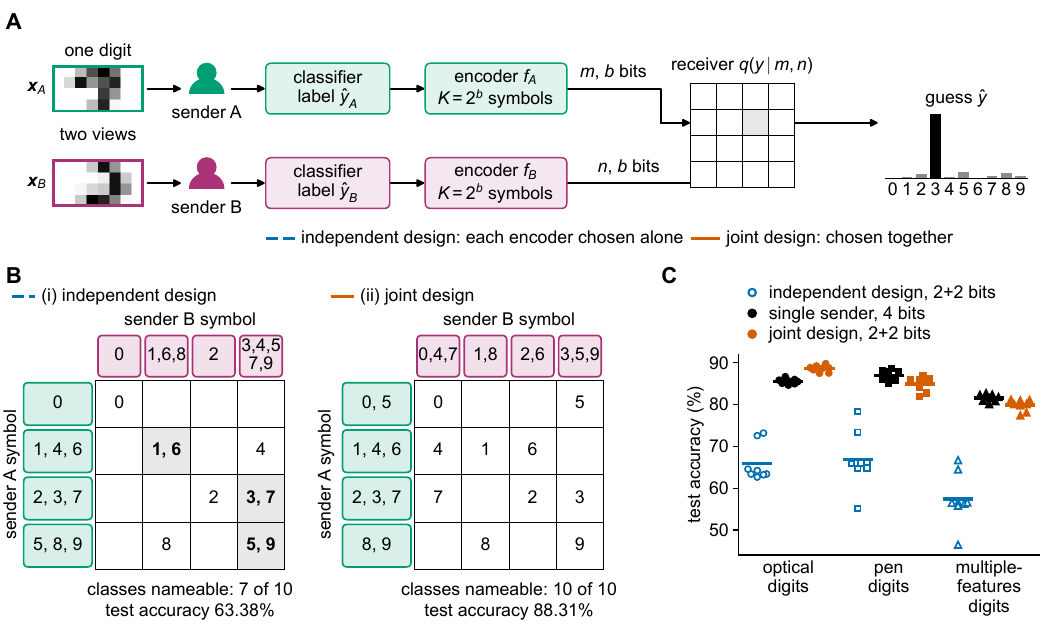}
\caption{\textbf{Two senders, short messages and a learned receiver.} \textbf{A} A handwritten digit is split into a top view $\bm{x}_A$ and a bottom view $\bm{x}_B$. Each sender classifies its view, and its code ($f_A$ or $f_B$, labelled encoder in the panel) maps the predicted label to one of $K=2^b$ symbols. Sender A sends the symbol $m$ and sender B the symbol $n$. The pair selects one cell of the receiver's table $q(y\mid m,n)$, which holds class probabilities learned from labelled examples, and the receiver answers with the most probable class. The displayed probabilities are illustrative. Digit image adapted from Alpaydin and Kaynak's UCI Optical Recognition of Handwritten Digits dataset (CC BY 4.0), split into two views. \textbf{B} Real two-bit codes for one training split and one assignment of classifier types (logistic sender A, multilayer perceptron sender B) on optical digits~\citep{AlpaydinKaynak1998}. Labels above each grid show how each sender groups its predicted classes into symbols. Each grid entry is the class that lands in that cell when both senders predict correctly. Classes nameable is the maximum number of classes the receiver can answer correctly, one per distinct message pair, when both senders predict correctly. \textbf{(i)} Independently designed codes produce \nCtlCrossOptSeedZeroLMIndepTwoBitNameable{} distinct pairs. Three shaded cells each merge two digits, of which the receiver can name only one. Test accuracy is \nOptUnitSeedZeroLMPairIndepAcc\%. \textbf{(ii)} Jointly designed codes give every digit its own pair, with test accuracy \nOptUnitSeedZeroLMPairJointAcc\%. \textbf{C} Test accuracy at four total bits for independently designed 2+2 codes, the whole-budget solo and jointly designed 2+2 codes (legend). Markers show eight units per dataset (two assignments of classifier types to views, four training splits); horizontal lines, their means. Circles, optical digits; squares, pen digits~\citep{AlpaydinAlimoglu1996}; triangles, multiple-features digits~\citep{Duin1998}. On optical and pen digits the solo sends the multilayer perceptron's label through a fitted receiver. On multiple-features digits it is the classifier trained on Zernike moments, using its own decision, a choice made after the results were seen (SI Appendix, section~S13). The units of a dataset share its test objects (\nOptObjTest, \nPenObjTest{} and \nMfeatObjTest{} objects, respectively).}
\label{fig:m-game}
\end{figure}

Picture a handwritten digit cut into a top half and a bottom half. One classifier sees only the top, another only the bottom. Each predicts the digit and sends a short message to a third party, the receiver, which never sees the image and must name the digit from the two messages alone (Fig.~\figref{fig:m-game}{A}). We call the classifiers senders, and the senders with their receiver a team. Each view is read by one of two classifier types, a logistic regression or a small neural network, a multilayer perceptron. Swapping the two classifier types between the views and repeating the analysis on four random training splits gives eight experimental units per dataset. Each unit is one classifier-to-view assignment and training split, with its own fitted team. A sender cannot always transmit its full prediction. There are ten digit labels, but a two-bit message has only four values, so each sender must group several labels under one message. Its code, also called its encoder, is the rule that does the grouping. The receiver is a table with one cell for each pair of messages. From labelled examples it learns which classes land in each cell, and on a new object it answers with the most probable class (SI Appendix, sections~S1 and~S2).

The two codes can be designed apart or together. In independent design, each sender's code is chosen as if that sender were the receiver's only source. In joint design, the two codes are chosen for how well the receiver can use them together. Fig.~\figref{fig:m-game}{B} shows real two-bit codes of one pair of senders. To isolate the effect of the codes, suppose both senders predict correctly. Classes that produce the same message pair are indistinguishable to the receiver, which must choose one answer for that pair. The number of distinct pairs therefore gives the maximum number of classes for which the receiver can consistently give the correct answer. Designed apart, the codes produce \nCtlCrossOptSeedZeroLMIndepTwoBitNameable{} distinct pairs (Fig.~\figref{fig:m-game}{B(i)}). Designed together, they give all ten classes separate pairs (Fig.~\figref{fig:m-game}{B(ii)}). The senders are the same. Only the words changed. We call the accuracy gained by joint over independent design, at the same division of bits, the coordination gain.

A second sender takes bits that the first could have used alone. We therefore also give the whole budget to one sender and keep its partner silent. This is the whole-budget solo, or simply the solo. With four bits, one sender can send any of the ten labels without grouping them. The team's accuracy minus the solo's accuracy, at the same total budget, is the cooperation gain. We are particularly interested in whether cooperation helps when both senders already classify accurately on their own. We call a sender capable when it reaches 80\% accuracy, the competence threshold, on selection data, development objects not used to train it or to fit its messages, and we ask whether two capable senders can outperform the solo with the same total number of bits. Fig.~\figref{fig:m-game}{C} places the three designs side by side at four total bits on three digit datasets: optical digits~\citep{AlpaydinKaynak1998} (the top and bottom halves of a small digit image), pen digits~\citep{AlpaydinAlimoglu1996} (the horizontal and the vertical coordinates of a pen trajectory) and multiple-features digits~\citep{Duin1998} (two numerical descriptions of one image). Both senders are capable on optical digits (SI Appendix, section~S2.4). On pen and multiple-features digits one sender falls below the threshold (SI Appendix, section~S2.1). Teams designed apart lose to the solo everywhere. Teams designed together beat it on one dataset and trail it on two. Coordination asks whether designing the messages together helps the team; cooperation asks whether the team beats one sender given the whole budget. The two questions have different answers. Data splits, the fitted receiver, the design objective and the selection rules are given in full in SI Appendix, section~S2.

\subsection{Designing the codes together}\label{sec:m-words}

\begin{figure}[!p]
\centering
\includegraphics[width=\textwidth]{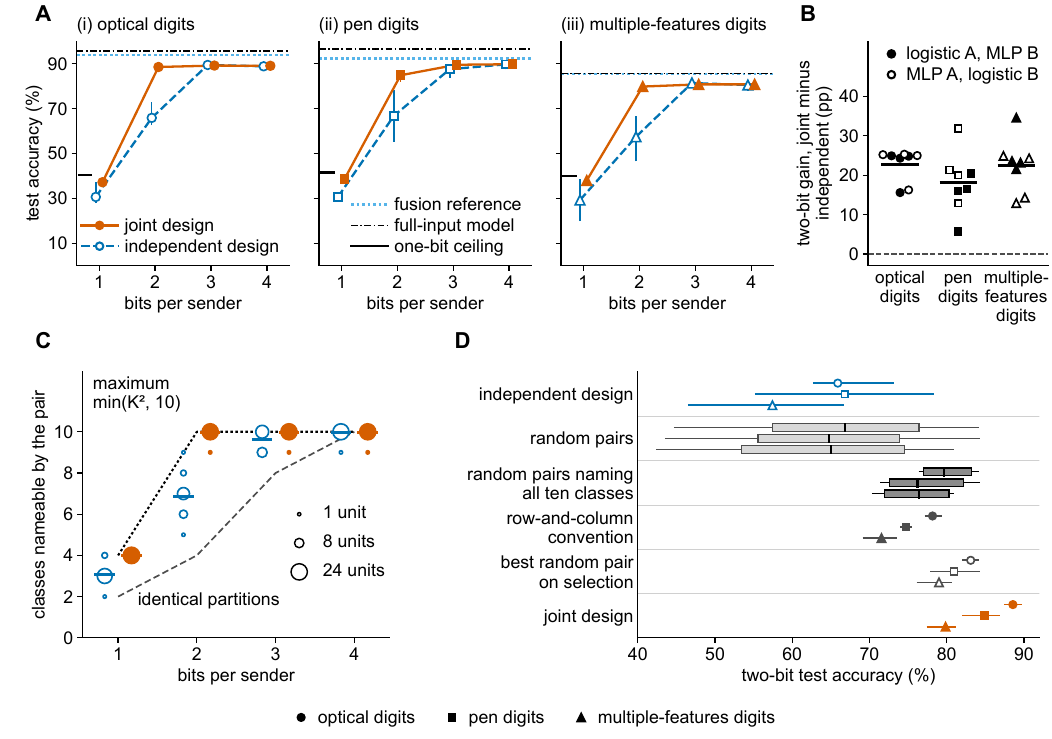}
\caption{\textbf{Designing the codes together lets two short messages name every class.} Class-label messages of $b$ bits ($K=2^b$ symbols) per sender. Joint design fits the two codes together. Independent design fits each code as if its sender were the receiver's only source. In A, C and D these designs use vermilion filled and blue open markers, respectively. In A their lines are solid and dashed. Dataset shapes in A, B and D identify optical (circles)~\citep{AlpaydinKaynak1998}, pen (squares)~\citep{AlpaydinAlimoglu1996} and multiple-features digits (triangles)~\citep{Duin1998}. \textbf{A} Test accuracy against bits per sender for \textbf{(i)} optical, \textbf{(ii)} pen and \textbf{(iii)} multiple-features digits. Markers are means over eight units, with lowest-to-highest bars. The legends in (i) and (ii) apply to all three. The fusion reference multiplies the two senders' full class-probability vectors with no message budget. The full-input model reads the whole input. Four possible message pairs allow at most four predicted classes at one bit per sender. The short black bar is their accuracy ceiling, the combined frequency of the four most common test classes. \textbf{B} Coordination gain at two bits per sender in each of the 24 units (filled, logistic sender on view A; open, multilayer perceptron (MLP) on view A); horizontal lines, dataset means; dashed line, zero. \textbf{C} Partition crossing, the number of classes the two codes can name when both senders are correct, against bits per sender, pooled over the 24 units (circle area, number of units; dashes, means); dotted line, $\min(K^2,10)$ for crossed groupings; dashed line, $\min(K,10)$ for identical groupings. \textbf{D} Two-bit test accuracy of six ways of building the codes, one row per dataset within each scheme, from top to bottom: designed apart (mean, lowest-to-highest bar); the same \nCtlRandDraws{} random code pairs in every unit, pooled over eight units (light boxes), and the \nCtlRandInjectiveDraws{} of them that name all ten classes (darker boxes); the row-and-column convention; the best random pair chosen on selection data; and designed together. Boxes span the 5th to 95th percentiles, with black median ticks and whiskers at the extremes. Filled and open dark-grey markers identify the convention and selection-chosen random pair. For the four data-blind schemes only the receiver is fitted.}
\label{fig:m-words}
\end{figure}

With the trained classifiers fixed and two bits per sender, joint design raises test accuracy over independent design by \nOptGainTwoBitPPAbs, \nPenGainTwoBitPPAbs{} and \nMfeatGainTwoBitPPAbs{} percentage points (pp) on optical, pen and multiple-features digits (Fig.~\figref{fig:m-words}{A}). The gain is positive in all \nPoolHardPairTwoBitNPos{} experimental units (eight per dataset) (Fig.~\figref{fig:m-words}{B}). Joint design also reduces the communication needed to approach full-label performance. A jointly designed pair sending two bits each comes within a fraction of a point of two senders sending their full labels at four bits each on two of the three datasets (SI Appendix, section~S5).

Counting the distinct message pairs explains why joint design can help. A sender with four symbols divides the ten classes into four groups. If both senders form the same groups, the receiver can tell apart at most four classes. If their groups cross, the pair has sixteen message combinations, enough for all ten classes. Designed apart, each sender spends its symbols on the distinctions that serve it best, and those overlap. At two bits the independent codes name at most \nCtlCrossAllIndependentTwoBitDistinctSymbolPairsMax{} classes, whereas the joint codes name all ten in \nCtlCrossAllJointTwoBitNUnitsNameAllTen{} of 24 units (Fig.~\figref{fig:m-words}{C}). The counting generalizes. Suppose each of $n$ senders has $K$ symbols and there are $C$ classes. Crossed groupings can name every class while repeated groupings cannot, exactly when $K<C\le K^n$. We call these budgets the coordination window. For ten equally frequent classes known exactly to both senders, the ideal accuracy gap between complementary and identical groupings peaks at two bits per sender, the fewest bits inside this window (SI Appendix, sections~S8 and~S4). Inside the window, crossing is needed to name every class, but independently designed codes may cross by accident. At three bits they already name nine or ten classes by themselves, and the gain falls to between \nMfeatHardPairThreeBitGainPP{} and \nPenHardPairThreeBitGainPP\pp. At one bit no pair of codes can name more than four classes (Fig.~\figref{fig:m-words}{A}). In two subsequent studies, we predicted before testing that the gain would peak at the first team size inside this window for nested teams of up to six digit views and teams of three to five body-worn sensors~\citep{UCIdsa}, and it did (SI Appendix, sections~S16 and~S17).

We next ask how much of the gain a fixed convention can recover without fitting the encoders. Arrange the ten labels in rows of four. Let one sender report the column of its predicted label and the other the row. When both are right, row and column identify the digit. This agreement is fixed before any data are seen, and only the receiver is learned. It recovers \nCtlConvOptTwoBitConventionFracJointGainRecoveredPct, \nCtlConvPenTwoBitConventionFracJointGainRecoveredPct{} and \nCtlConvMfeatTwoBitConventionFracJointGainRecoveredPct\% of the coordination gain on the three datasets (Fig.~\figref{fig:m-words}{D}). Picking the best of \nCtlRandDraws{} random code pairs on selection data closes part of the remaining gap, although only \nCtlRandInjectiveDraws{} of those pairs name all ten classes (Fig.~\figref{fig:m-words}{D}). In a comparison with predefined complementary codes, joint design retained an accuracy advantage of \nRevMfeatCommonGain{} to \nRevPenCommonGain\pp\ over the best, on selection data, of \nRevLibrarySize{} fixed code pairs that were all built to name ten classes (SI Appendix, sections~S6 and~S6.3). Agree on who says what before learning anything, then learn the details.

To test whether coordination gains depend on correlations beyond the shared class, we rebuilt the data so that the views were independent within each class while preserving each sender's own statistics. The gains changed by \nOptGainNullMinusNatPP, \nPenGainNullMinusNatPP{} and \nMfeatGainNullMinusNatPP\pp, each less than one point in magnitude, showing that the benefit of joint code design persisted without those correlations (SI Appendix, section~S7).

\subsection{Who should speak is a property of the pair}\label{sec:m-pair}

A jointly designed team can still perform worse than the whole-budget solo. Two senders with two bits each have sixteen possible message pairs, and one sender with four bits has sixteen possible messages. The counting that explains coordination gives a team no advantage in capacity over a solo with the same budget (SI Appendix, section~S8). A second sender must bring evidence that the first lacks, the messages must preserve that evidence, and the receiver must learn to use it. Each requirement can fail on its own.

\begin{figure}[!p]
\centering
\includegraphics[width=\textwidth]{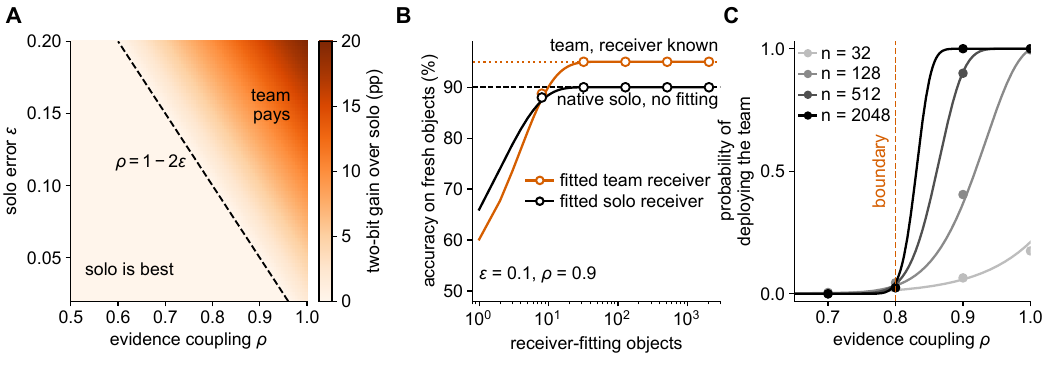}
\caption{\textbf{Who should speak is a property of the pair.} \textbf{A} The largest gain over the whole-budget solo that any two-bit team can achieve, $G_2^{\ast}$ of Eq.~\eqref{eq:m-frontier}, in percentage points (colour), against the coupling $\rho$ of the two senders' private bits and the solo error $\epsilon$. Along every horizontal line both senders keep the same individual statistics and solo accuracy $1-\epsilon$. Left of the dashed boundary $\rho=1-2\epsilon$ no team can help. \textbf{B} Accuracy on fresh objects against the number of receiver-fitting objects in the same family at $\epsilon=0.1$ and $\rho=0.9$. Solid curves give exact expectations for the fitted receivers, and open markers show 200-simulation means (legend). The original-prediction solo requires no receiver fitting (black dashed, $1-\epsilon$). The team with its receiver known scores $(1+\rho)/2$ (vermilion dotted). \textbf{C} Detecting a positive gain at $\epsilon=0.1$: the probability that a one-sided paired binomial test at the 5\% level approves the team with its receiver known over the solo using its classifier's original prediction, against $\rho$, for \nRegNatSmallCalibrationN, 128, \nRegNatLargeCalibrationN{} and 2,048 calibration objects (legend). Curves give exact probabilities; points, frequencies over 200 synthetic repetitions. The dashed vermilion line marks $\rho=1-2\epsilon$.}
\label{fig:m-pair}
\end{figure}

We ask whether each sender's individual statistics determine the value of pairing them. The following construction shows that they do not. Consider a yes-or-no question with equally likely answers and a total budget of two bits. Two senders both see a shared cue that is correct with probability $1-\epsilon$, where $0\le\epsilon\le1/2$, and each also sees one fair private bit. Alone, a sender can do no better than report the cue, so the whole-budget solo errs with probability $\epsilon$. A sender's accuracy, the reliability of its confidence and the whole relation between its observation and the answer are all fixed by $\epsilon$. What varies is how the two private bits relate. With probability $\rho\in[0,1]$, we generate the two private bits so that they differ when the cue is wrong and match when it is correct. Their exclusive-or ($\oplus$)---one for different bits and zero for matching bits---then indicates whether the cue is wrong. With the remaining probability $1-\rho$, the bits are drawn independently. The largest gain over the solo that any codes, any division of the two bits and any receiver can achieve is
\begin{equation}\label{eq:m-frontier}
G_2^{\ast}(\epsilon,\rho)=\tfrac12\,[\rho-(1-2\epsilon)]_+ ,
\end{equation}
where the subscript marks the two-bit budget and $[x]_+$ keeps the positive part of $x$ (SI Appendix, section~S9). Below the boundary $\rho=1-2\epsilon$ no team beats the solo. Above it, one sender takes the exclusive-or of its cue and its private bit, the other sends its private bit, and the receiver's exclusive-or recovers the answer whenever the bits are coupled. At $\rho=1$ the pair removes every error the solo makes (Fig.~\figref{fig:m-pair}{A}). The mechanism is the exclusive-or that defines synergy in information decompositions~\citep{Williams2010,Schneidman2003}. Equation~\eqref{eq:m-frontier} gives the largest achievable team gain under a fixed total bit budget. As $\rho$ varies at fixed $\epsilon$, the individual statistics stay unchanged, including accuracy and calibration. Both senders remain capable when $\epsilon\le0.2$. Although each sender's individual performance stays unchanged, the relationship between their private bits varies. When that relationship is insufficiently informative, no team improves on the solo. When it reveals every error in the shared cue, the team corrects them all. Measurements of either sender alone cannot reveal the relationship between their private bits, or whether that relationship allows the team to correct the shared cue's errors.

The evidence that makes a partner valuable can be invisible in each partner's calibrated judgement. In this construction, the private bits can reveal errors in the shared cue only when combined. Each sender's exact class probabilities omit its private bit, so probability reports cannot improve on the solo, regardless of their length. The same loss of information appears in a constructed text task. Each sender reads a description of one bit, and the receiver must decide whether the two bits agree. Neither bit alone predicts the answer, but together they determine it. Messages formed from internal language-model states preserved enough information to solve the task almost perfectly, whereas messages based on predicted classes or class probabilities remained near chance (SI Appendix, section~S14).

If the views are independent once the class is known, however, their exact class probabilities suffice for optimal message design at the same bit budget~\citep{Xu2012,XuZhuChen2013} (SI Appendix, section~S3). This conditional-independence assumption is also used in classical co-training, which learns from two views~\citep{Blum1998}. Preserving useful evidence and learning how to combine it are separate requirements. Even when messages preserve useful joint evidence, separate measurements of the senders may not establish whether a particular team helps or hurts (SI Appendix, section~S12). Calibrate in pairs---estimate what each message pair means from labelled objects processed by both senders. Paired data can reveal how to combine the messages, but cannot recover information they have discarded.

Suppose the evidence exists and the messages preserve it. The receiver must still learn what each pair of messages means from a finite number of paired examples. In the family above at $\epsilon=0.1$ and $\rho=0.9$, where the team's true advantage is five points, a receiver fitted on \nRegNatSmallCalibrationN{} paired objects already reaches \nNatSmallReceiverAccuracy\% accuracy, close to a receiver that knows the true cell probabilities (Fig.~\figref{fig:m-pair}{B}). In this example, relatively few observations suffice to fit the receiver, but many more are needed to establish its advantage statistically. A paired test that must show the team beating the solo before switching it on, applied to the team with its receiver known, approves deployment in \nNatSmallActivation\% of cases with \nRegNatSmallCalibrationN{} calibration objects and in \nNatLargeActivation\% with \nRegNatLargeCalibrationN{} (Fig.~\figref{fig:m-pair}{C}). The receiver's exact learning curve, and a label-free check that a team's messages can reproduce the decisions of the solo already in use, are given in SI Appendix, sections~S10 and~S11.

\subsection{When the second sender pays}\label{sec:m-map}

\begin{figure}[!p]
\centering
\includegraphics[width=\textwidth]{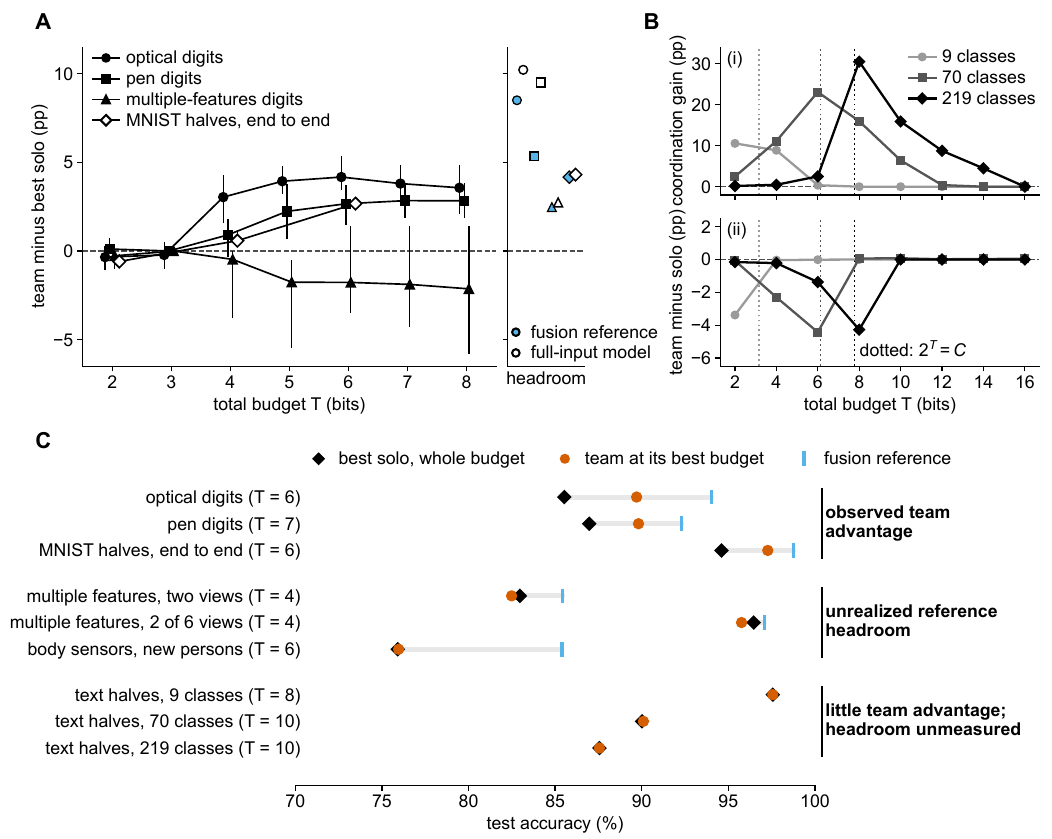}
\caption{\textbf{When the second sender pays: three outcomes on natural data.} Gains above zero in A and B(ii) mean that the team beats one sender given the whole budget. The solo receives all $T$ bits per object. In A (except MNIST) and B(ii), the team chooses its division of $T$ on selection data. \textbf{A} Team minus solo, in percentage points of test accuracy, against $T$. For optical (circles)~\citep{AlpaydinKaynak1998} and pen digits (squares)~\citep{AlpaydinAlimoglu1996} the solo uses the stronger sender's label code. It carries the label without loss from four bits, and its accuracy then stays constant. For multiple-features digits (triangles)~\citep{Duin1998} the solo is the Zernike sender's label code, held at its lossless four-bit value from five bits on. MNIST halves (open diamonds)~\citep{LeCun1998} use a team with equal bits per sender and the strongest tested solo, including encoders trained end to end, over \nEeSeeds{} training splits. Points are means, with lowest-to-highest bars over units or splits. The separate column shows gains over the solo at its full label for the fusion reference and one model reading the whole input (legend). \textbf{B} Text halves (DBpedia; \citep{DBpediaHierarchy}; means over 16 ordered language-model pairs), with class counts identified in the legend, against $T$: \textbf{(i)} coordination gain with equal bits per sender; \textbf{(ii)} the team with its division of $T$ chosen on selection data, minus the solo. Dotted verticals mark $T=\log_2C$, beyond which the solo's label is lossless. \textbf{C} Every studied team on one strip of test accuracy, showing the solo, team at its best budget and fusion reference (legend; budget in the row label, solo at the same budget; fusion undefined for text). Grey bars span the solo to the fusion reference. Rows are grouped by outcome, named at the right. The six-view row shows the best two of the six multiple-features views at two bits each. The three-sensor team from Daily and Sports Activities~\citep{UCIdsa}, at two bits per sensor, failed the 80\% competence requirement (only the left-leg sensor cleared it on development data) and is evaluated on recordings from four participants whose data were not used to fit or select the team. Group labels describe measured outcomes after the fact and do not identify the cause of a shortfall (SI Appendix, section~S13).}
\label{fig:m-map}
\end{figure}

On natural data the question becomes empirical. We compare each team with the solo and, where available, with a fusion reference, the product of the two senders' full class probabilities, which uses both views with no message budget. The accuracy difference between the fusion reference and the solo shows an improvement achieved by combining the uncompressed outputs. The team--solo difference measures the improvement achieved under the communication budget. We distinguish three outcomes: the team improves on the solo; it provides little or no improvement despite an advantage achieved by the fusion reference; or it provides little or no improvement and no fusion reference is available (Fig.~\figref{fig:m-map}{}).

Once the solo can transmit its full predicted label, additional bits cannot improve that class-label message. The team can use those bits to report the other view. On optical digits the team leads the solo in every unit from four total bits on, and on pen digits from five, although there the weaker sender itself falls below the 80\% threshold (SI Appendix, section~S2.1). At six bits the second sender removes \nRegOptTSixErrRed{} and \nRegPenTSixErrRed\% of the solo's errors on the two datasets (Fig.~\figref{fig:m-map}{A}). On MNIST~\citep{LeCun1998}, the standard handwritten-digit dataset, we also compare the team with single senders whose image representations and message encoders are learned together to help the receiver classify the digit, using the same total communication budget. The two senders are neural networks trained together with the receiver, through the bit limit, on the top and bottom halves of each image, and they face the strongest single sender we tested with the same budget. Six shared bits raise accuracy by \nEeDiffJointEncMinusSoloBThreeMean\pp, with a paired image-resampling interval of $+\nCiJeSoloSixLower$ to $+\nCiJeSoloSixUpper$\pp, and remove \nEeErrRedJointEncVsSoloBThree\% of that solo's errors on average, with the team ahead in all \nEeSeeds{} training splits (SI Appendix, sections~S13 and~S18).

For the same MNIST halves, we compare codes that compress the frozen classifiers' predicted labels with neural encoders trained jointly with the receiver. At four total bits---two per sender---the class-label codes achieve a mean test accuracy of \nEeJointLabelBTwoAcc\%, compared with \nEeJointEncBTwoAcc\% for neural encoders trained from scratch, an advantage of \nEeDiffJointEncMinusJointLabelBTwoMeanAbs{} percentage points. At six total bits, both reach \nEeJointEncBThreeAcc\%. In an exploratory comparison, initializing the neural encoders from trained classifiers before training them with the receiver raises their six-bit accuracy to \nEeWsJointEncBThreeAcc\% (SI Appendix, section~S18).

On optical, pen and multiple-features digits, the team also chooses how to divide its bits on selection data. On pen digits this choice gives three bits to the stronger sender and one to the weaker in all eight units, which turns the equal-split loss of Fig.~\figref{fig:m-game}{C} into a mean gain of \nRegPenTFourDiff\pp. On multiple-features digits it silences the weaker sender in \nCtlZernMfeatHardTFourNFavoursZernike{} of 8 units, so that the team becomes a solo (SI Appendix, sections~S5.7 and~S13.6). A team should be allowed to become a solo.

A large coordination gain does not guarantee a team advantage. Encyclopedia abstracts~\citep{DBpediaHierarchy}, labelled by topic at three levels of detail (9, 70 and 219 classes), were cut into two halves of their words. Each half was read by a pretrained language model whose representation was left unchanged, with a small classifier trained on top. With equal bits per sender, coordination gains peak at \nLlmSeventyCoord\pp\ for 70 classes and \nLlmFineCoord\pp\ for 219 classes (Fig.~\figref{fig:m-map}{B(i)}). Yet the team that chooses its division of bits on selection data never beats the solo by more than \nRegLlmMaxGainAllLevels\pp\ for any language-model pair, level or budget, and at 70 classes and six total bits it still trails (Fig.~\figref{fig:m-map}{B(ii)}). Splitting those six bits equally would have done far worse: the equal-split team's accuracy minus the solo's is $\nLlmSeventySoloGap$\pp. This analysis did not include a fusion reference, so it could not establish whether the partner lacked useful evidence or whether the messages failed to preserve it. We subsequently checked whether combining the senders' uncompressed outputs could outperform the solo on development data. Fusion did outperform the solo, showing useful complementary evidence that the budgeted teams did not exploit on those data (SI Appendix, sections~S13 and~S20).

On 219-class text, transmitting both predicted labels in full still left the fitted receiver below the solo (\nNatIdentityThreeRawAccuracy\% against \nLlmHiddenFineLabelSolo{}\%), although it could reproduce that solo simply by ignoring the second message (SI Appendix, section~S15.3).

We extended the multiple-features digit analysis to teams of two to six classifiers, each reading a different numerical description of the same digit. Designing their codes together produced large coordination gains over independently designed codes, but no team outperformed the best single-view classifier acting alone. Joint design reaches the counting ceiling exactly, naming all ten classes wherever $K^n\ge10$, and the four-sender one-bit team gains \nNsGainMaxAbs\pp\ over independent design. The best single view alone reaches \nNsTopView\%, within two points of the fusion reference (SI Appendix, section~S16).

Fig.~\figref{fig:m-map}{C} sorts every studied team into the three outcomes, beside its solo and its fusion reference. The team's mean test accuracy exceeds the solo's on optical digits, pen digits and MNIST halves. On the others it trails, or ties within a tenth of a point, including the body sensors of the next section.

\subsection{Deciding before deployment}\label{sec:m-decide}

An engineer wants the answer before deployment. We use labelled examples to predict the team's advantage on held-out observations. We call these preliminary measurements calibration. Separate calibration estimates how each sender's messages relate to the true class, without estimating how the two senders' messages occur together. Paired calibration measures both senders on the same objects. The solvable model of Section~\ref{sec:m-pair} shows that the same separate measurements can describe pairs with different gains. Separate calibration can yield a correct prediction for a particular pair, but identifying which gain applies requires either an assumption about how the views relate or enough paired calibration objects. We examine this question on dry beans~\citep{DryBean2020}, body-worn sensors, paired bird~\citep{Wah2011CUB,Reed2016} and research-paper~\citep{WOSdata} views, and encyclopedia text. In the text analysis, one prediction rule combined separate sender measurements by assuming that their predictions were independent given the true class. This tested whether separate measurements could suffice to predict cooperation. The same analysis also used paired observations to fit receivers and evaluate another selection rule.

\begin{figure}[!p]
\centering
\includegraphics[width=\textwidth]{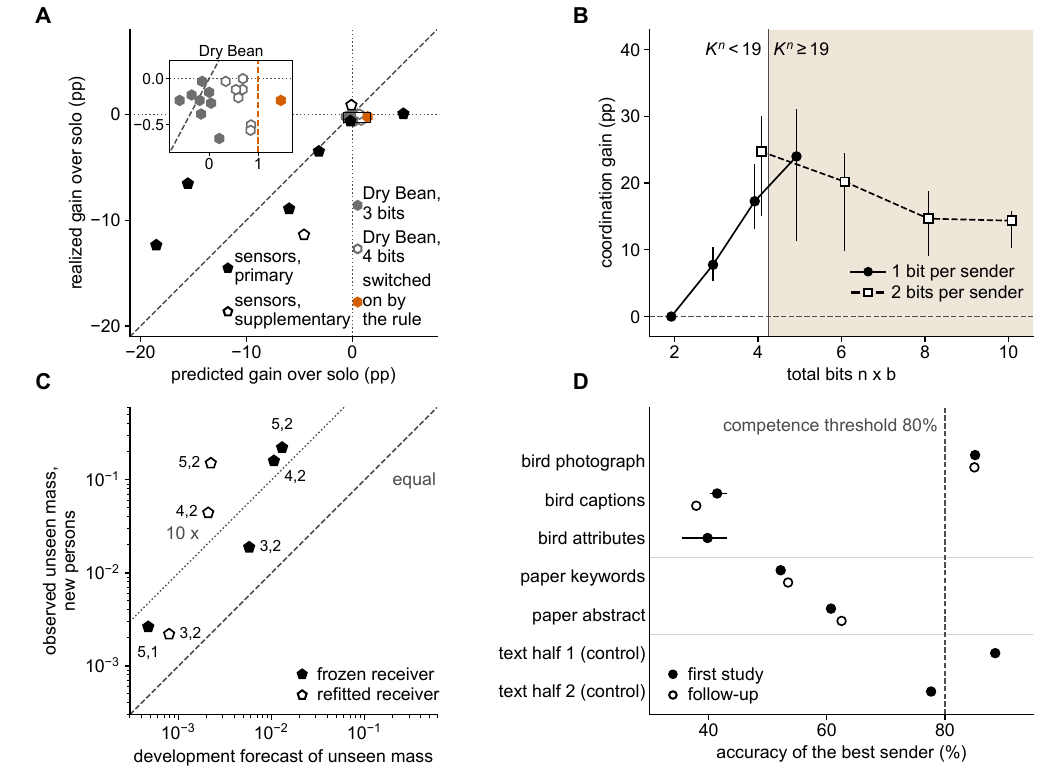}
\caption{\textbf{Deciding before deployment: what transferred to new objects and new people.} \textbf{A} Gain of the team over the whole-budget solo predicted before testing (horizontal) against the realized gain (vertical), in percentage points. Pentagons show body-worn sensors from Daily and Sports Activities~\citep{UCIdsa}, predicted from paired calibration on development persons and realized as means over four held-out persons with codes and receivers frozen (filled, three to five sensors; open, two sensors). Hexagons show dry-bean classification~\citep{DryBean2020}, with one sender reading size measurements and the other shape measurements, predicted from a joint table assembled from separate calibration cohorts (filled grey, three total bits; open grey, four; vermilion, the one design the frozen rule switched on). The inset enlarges the Dry Bean designs near the origin. Its vermilion dashed line marks the activation threshold of one predicted percentage point. Dashed diagonals mark equality. Dotted lines in the main panel mark zero. \textbf{B} Coordination gain on the held-out persons against total bits $n\times b$ for two to five sensors (filled circles, one bit per sensor; open squares, two bits), means with lowest-to-highest bars over persons. The vertical line marks $K^n=\nSenClasses$, the start of the coordination window (shaded). The prediction for teams of three to five sensors, recorded before testing, was that the one-bit series peaks at five sensors, the first one-bit team inside the window. \textbf{C} Unseen message combinations are combinations of sensor messages that never occurred together during receiver fitting. The horizontal axis shows the predicted fraction of activity recordings producing these combinations, estimated from development data; the vertical axis shows the observed fraction, averaged over four new participants. Both axes are logarithmic ($0.01=1\%$). Filled symbols: codes and receiver learned from one person's recordings and applied unchanged to four other people. Open symbols: each of those four people is evaluated using codes and a receiver learned from the other three people's recordings. Labels give the number of sensors and bits per sensor. Dashed line: observed equals predicted; dotted line: observed is ten times predicted. Only conditions with nonzero observed fractions are shown (7 of 16). \textbf{D} Development accuracy of the best sender for each bird~\citep{Wah2011CUB,Reed2016} and research-paper~\citep{WOSdata} view (dashed line: \nCvGateThreshold\% competence threshold). Filled markers show selection accuracies in the first study, with bars spanning five split seeds. Open markers show qualification accuracies after source and readout selection on a separate subset of the same development cohort. The follow-up used a quantized seven-billion-parameter language model for captions, keywords and abstracts, and re-measured the photograph sender. Bird attributes, a third sender reading the archive's binary attribute annotations and used only in three-sender teams, do not enter the pair requirement. The two text halves from DBpedia~\citep{DBpediaHierarchy} show the procedure on a task whose outcome is known: one clears the threshold and the other falls just short. No pair reached it.}
\label{fig:m-decide}
\end{figure}

In the dry-bean classification task, one sender read size measurements and the other shape measurements extracted from photographs. We predicted the team's performance from separate measurements of the two senders, assuming their predictions were independent once the true class was known. Eight fitted sender configurations were evaluated at three and four total bits, giving sixteen cases. The rule selected a two-sender team only if its predicted accuracy exceeded the solo's by more than one percentage point. Otherwise, it assigned the whole budget to one sender. The rule retained the solo in fifteen of sixteen cases, consistent with the observed performance of the rejected teams. In several of these cases, the rule predicted a small positive gain even though the team lost on test. Its one-point threshold nevertheless kept the solo in use, giving a correct decision despite an incorrect prediction of the gain's sign. The only partnership it approved did not show the predicted improvement. In that four-bit case, the predicted gain was \nRevBeanActivePredicted\pp, whereas the observed change was \nRevBeanActiveGain\pp, with a resampling interval of \nRevBeanActiveLower{} to \nRevBeanActiveUpper\pp (Fig.~\figref{fig:m-decide}{A}, inset; SI Appendix, section~S19).

On 219-class text, predicting joint performance from separate sender measurements led the rule to select \nLlmFineQCount{} teams, yet every selected team performed worse than the whole-budget solo. The mean predicted gain was $+\nLlmFineQPred$\pp, whereas the mean observed change was $\nLlmFineQObserved$\pp (SI Appendix, section~S14). Separate measurements therefore failed to predict which partnerships would improve accuracy in this task. This illustrates why knowing how each sender performs does not, by itself, establish whether sharing the channel will help.

On body sensors, predictions from paired calibration on four development participants were recorded before testing. For all \nSenPTwoTOnePrimaryCells{} comparisons involving three to five sensors, we correctly predicted whether the team's average accuracy across the four new participants would be higher or lower than the solo's. \nSenPTwoTOnePrimaryPredLossCells{} comparisons showed a loss, and the remaining gain was very small. Only \nSenGateAdmissible{} of the \nSenUnits{} sensor classifiers reached the 80\% threshold. A rule that always predicts a loss would have matched five, and the one predicted benefit, \nSenDevBTwoNThreePTwoPoint\pp, realized \nSenTOneBTwoNThreeGainSolo\pp\ (Fig.~\figref{fig:m-decide}{A}; SI Appendix, section~S17).

What did transfer to new people was coordination. The prediction that, for teams of three to five sensors, joint design would gain most where five one-bit sensors first name all \nSenClasses{} activities held, with a gain of \nSenPFourTOneArgmaxGainAbs\pp\ on the new persons. A supplementary two-sensor, two-bit design just below the window gained \nSenTOneBTwoNTwoGainIndep\pp\ (Fig.~\figref{fig:m-decide}{B}). Counting travels: among teams of three to five sensors, the predicted peak in coordination gain appeared on new participants. What travelled less well was the receiver's ability to interpret the messages: many message combinations on these participants had not occurred during receiver fitting.

Unseen message combinations were common on new persons. A combination never seen during fitting leaves the receiver guessing from the class frequencies alone, and its expected share can be computed in advance from the fitting data. In the \nSenPThreeACellsWithMass{} combinations of design and fitting rule, out of 16, where any occurred, the observed share exceeded the forecast by factors of \nSenPThreeARatioMin{} to \nSenPThreeARatioMax, and in the worst design \nSenTeamTOneUnseenMassMax\% of a new person's segments produced a combination the receiver had never seen (Fig.~\figref{fig:m-decide}{C}; SI Appendix, sections~S10 and~S17). Population change, dependence among a person's segments and the fitted codes may each contribute, and their shares could not be separated. Receiver fitting must account for variation in the population where the system will be used.

The bird and research-paper studies first checked whether both senders met the competence threshold, before evaluating the held-out test data. We paired a photograph of a bird with the captions written for it (\nCvCubClasses{} species) and the keywords of a research paper with its abstract (\nCvWosClasses{} research areas), and required both senders of a pair to reach \nCvGateThreshold\% on selection data before any test. The photograph sender reached \nCvCubImageAcc\%, the best caption sender \nCvCubCaptionAcc\%, keywords \nCvWosKeywordAcc\% and abstracts \nCvWosAbstractAcc\% (Fig.~\figref{fig:m-decide}{D}). A follow-up that read the captions, keywords and abstracts through a quantized language model of about seven billion parameters, on a separate qualification subset of the existing development cohort, left those senders far below the threshold (Fig.~\figref{fig:m-decide}{D}, open markers). On the first study's development data at the counting-window budgets reported here, coordination gains were large, yet the teams remained below their stronger whole-budget solos. On birds even the fusion reference scored below the photograph alone, and on research papers the fusion reference stood above the abstract alone, headroom that the budgeted teams did not realize (SI Appendix, section~S20). Neither pair met the competence requirement, so the held-out data were not evaluated.

\subsection{What to preserve when the question comes later}\label{sec:m-products}

The same coordination question arises when the receiver must estimate a product of features held by independent senders, neither of whom can predict the product alone. Here an independently drawn query tells the receiver which product is wanted only after the messages are fixed (SI Appendix, section~S21). For independent standard Gaussian sources and one real-valued linear message per sender, a constructed query distribution makes the uniquely optimal marginal directions---each chosen assuming the partner sends its full source---capture zero target variance together, whereas a joint choice captures $1/4$ of the unit target variance. 

More generally, the cost of choosing the summaries separately is bounded. Let $D_{\rm marginal}$ be the smallest mean squared error the receiver can achieve when each sender chooses its summary to minimize error assuming that the partner's complete measurements will be available. Let $D_{\rm joint}^{*}$ be the smallest error achievable by choosing both summaries together, under the same limits on the number of independent linear measurements each sender can transmit---the linear rank constraints. For independent standard Gaussian sources, with questions chosen independently of the measurements and finite target variance, we prove $D_{\rm marginal}\leq 2D_{\rm joint}^{*}$. Thus, separate choices can discard information that coordinated choices preserve, but their remaining mean squared error cannot exceed twice the joint optimum (SI Appendix, section~S22).

Choosing these linear summaries is mathematically a tensor-approximation problem: preserving the combinations of measurements that matter for the receiver's questions~\citep{DeLathauwer2000a,DeLathauwer2000b}. Messages restricted to one bit require a separate analysis. Here, scalar quantization---representing a continuous measurement by one of two symbols---provides the Gaussian bounds~\citep{Max1960,Lloyd1982} (SI Appendix, section~S23).

The preferred code also depends on the score. Let each sender hold three independent fair $\pm1$ coordinates and send one bit before a query is drawn uniformly from the nine coordinate products. Sending one coordinate minimizes mean squared error at $8/9$, whereas sending the majority sign maximizes sign accuracy at $5/8$ (SI Appendix, sections~S23 and~S24). As in random-access codes~\citep{Ambainis2008}, information must be compressed before the question is known. The distinctions a message should preserve depend on both the question distribution and the loss used to judge the answer.
\section{Discussion}\label{sec:discussion}

We asked when individually capable systems become jointly capable through a narrow channel, and when a second sender earns the bits it takes from the first. Who says what is a coordination problem: jointly designed codes give large gains, a fixed convention recovers much of the benefit without learning the codes, and counting identifies the coordination window. Whether to let a second sender speak at all is a cooperation problem whose answer depends on the relationship between the two observations. Its value must be measured against one sender given the whole communication budget.

Four rules for teams of learned systems follow. First, design the vocabulary together. A fixed complementary convention improves communication without fitting the encoders and recovers about half of the gain. Second, measure every team against the strongest single member given the whole budget, the comparison that multi-agent systems and human--machine teams often fail~\citep{Tran2026Single,Vaccaro2024,Smit2023MAD}. A coordination gain does not establish a cooperation gain: as the text halves show, improving the messages need not make the team better than the whole-budget solo. Third, estimate the receiver's class probabilities from examples processed by both senders. Identical individual statistics can accompany relationships that make partnership possible or impossible. Predictions based on measuring each sender separately failed in dry-bean and text classification, while measurements of the senders on the same objects predicted the signs on sensors, mostly of losses, without forecasting a useful benefit. A combination rule that sees only calibrated confidences cannot guarantee to beat the better partner~\citep{Peng2024NFL}, and the boundary $\rho=1-2\epsilon$ of Eq.~\eqref{eq:m-frontier} makes that limitation exact. Fourth, allow enough representative data to fit and evaluate the receiver. Count the message combinations that occur and estimate how often new combinations will appear in the intended population. Establishing an advantage statistically can require substantially more observations than fitting the receiver.

The results extend a geometric reading of the Babel story. Capable systems fail together when each writes the world in its own private coordinates, and whether a collective can hold a description in common is a question about translations around loops~\citep{Salahshour2026gauge}. Here we assume that the senders agree on the class meanings, and the receiver learns to interpret their codes. Even with shared class meanings, the senders' codes can preserve redundant distinctions. Their partitions determine which classes the message pairs can distinguish. Consistent translation decides whether senders can share a description at all. How their groupings of the classes cross decides which distinctions a shared description can carry through a narrow channel. The value of carrying them must still be measured against one sender.

Seen from evolutionary game theory, this is cooperation without a dilemma. Cooperation is usually studied against the temptation to defect~\citep{Nowak2006,SalahshourAltruistic2025}, and a shared code can emerge together with the cooperation it serves~\citep{Salahshour2020}. Here nobody is tempted, the members share one goal as in a team problem~\citep{MarschakRadner1972}, and the only cost of a partner is the share of the channel it takes. A partnership can still fail, for a structural reason. The second sender brings no evidence the first lacks, or the evidence does not survive the exchange, or the receiver cannot learn it from the examples at hand. The dilemma is replaced by a measurement, the team against the best single sender with the whole budget.

The evidence covers a specific range of models and tasks. Apart from the end-to-end encoders, the sources are frozen representations with supervised readouts: small classifiers on digits, dry beans and sensors, and readouts of four small open-weight language models, with a quantized seven-billion-parameter model in the bird and research-paper study. The language-model analyses concern class predictions from fixed representations, communicated through fitted message codes. Spreads over units, persons and training splits are descriptive. The grouping of datasets by outcome was made after the results were known. Most body-sensor classifiers did not meet the competence threshold, so their losses cannot be attributed to the receiver alone. See SI Appendix, section~S26 for further details.

The text control varied receiver fitting while keeping the senders and codes fixed. A next experiment could extend this comparison to MNIST, where the team improved on the solo, and to a task with two capable senders. To test cooperation between individually capable sources, we required both senders to meet the competence threshold before evaluating their team on held-out data. The quantized 7B follow-up did not bring the text senders above the required threshold.

A second sender earns its bits when the added evidence survives the exchange and the fitted team does better than the best tested use of those bits by one sender. Agreeing on meanings is a beginning. Deciding who should speak remains a question about the pair.
\section{Materials and Methods}\label{sec:mm}

\subsection{Datasets and Views} The two-view digit studies use optical digits~\citep{AlpaydinKaynak1998}, whose $8\times8$ pixel grids are split into top and bottom halves, pen digits~\citep{AlpaydinAlimoglu1996}, split into horizontal and vertical pen coordinates, and multiple-features digits~\citep{Duin1998}, with Fourier coefficients and Zernike moments as the two views and all six numerical descriptions in the six-view study. MNIST~\citep{LeCun1998} images are split into top and bottom halves, with test writers absent from training. Text halves are the two word halves of DBpedia abstracts labelled at 9, 70 and 219 nested classes~\citep{DBpediaHierarchy}. The body-sensor study uses the Daily and Sports Activities recordings~\citep{UCIdsa,Altun2010}: \nSenClasses{} activities performed by \nSenPersons{} persons and recorded by five inertial sensors, each summarized by simple statistics of its channels. Dry Bean~\citep{DryBean2020} supplies size and shape descriptors of dry beans. The complementary-views protocol pairs bird photographs with their captions~\citep{Wah2011CUB,Reed2016} and research-paper keywords with abstracts~\citep{Kowsari2017,WOSdata}.

\subsection{Senders} Each digit view is read by a multinomial logistic regression or a multilayer perceptron~\citep{Jain2000} with 64 hidden units, fitted on standardized features. Both assignments of the two classifier types to the two views are run on four random training splits, giving eight units per digit dataset. Dry Bean views are read by a logistic regression and an extremely randomized tree ensemble, and each body sensor by the better of the two digit classifier types on its development data. MNIST views are read by small convolutional networks, and in the end-to-end study by convolutional encoders trained through the message bottleneck together with the receiver. Training through these discrete messages uses Gumbel--Softmax~\citep{Jang2017,Maddison2017} and sign-based straight-through gradient estimators~\citep{Bengio2013}, compared in SI Appendix, section~S18. Text, caption, keyword and abstract senders are supervised linear readouts of frozen features from SmolLM2-135M, SmolLM2-1.7B, Qwen2.5-0.5B and Qwen2.5-1.5B~\citep{SmolLM2025,Qwen2025}, and, in a second run of the competence check, from a quantized Qwen2.5-7B. Bird photographs are read through DINOv2 features~\citep{Oquab2024DINOv2}. Language models supply fixed representations only. Model weights and tokenizers were obtained from the authors' repositories on the Hugging Face Hub. Exact repository identifiers and downloaded revisions are provided in the reproducibility package.

\subsection{Data Roles} Within each study the development objects are divided into representation training, which trains the classifiers, interface fitting, which fits the codes and the receiver, and selection, which chooses among message families, divisions of bits and solo comparators. The two-view digit studies use 60, 25 and 15 percent of the development objects for the three roles. The body-sensor study splits by person, the MNIST study uses test images from writers absent from training, and the complementary-views protocol uses the official bird split and a deterministic hash split of the research papers. The timing of test access, retrospective comparator choices and other exploratory analyses, and protocol deviations is summarized in the SI Appendix (section~S2.10).

\subsection{Messages and the Receiver} A sender first reduces its information to a discrete local value: its predicted label, or the label of one of 32 clusters of its class-probability vectors or internal feature vectors. Its encoder maps that local value to one of $2^b$ message symbols. The receiver holds one cell per combination of messages and stores the class counts of the interface-fitting objects that land there, smoothed by adding a total count of one, distributed across classes according to their overall frequencies. It predicts the most probable class. Codes are designed by minimizing the penalized Brier error~\citep{Brier1950} of the refitted receiver on the interface-fitting objects, moving one local value at a time to the symbol that lowers the objective most, from the independent codes and random starts, and retaining the code with the lowest objective value. Independent design uses the same move on each sender's own objective using a receiver that observes only that sender's message. The language-model study uses fitted accuracy as its search objective. Divisions of bits, message families and solo senders are chosen by selection accuracy, with ties broken by the selection Brier score. Full methods are given in SI Appendix, section~S2.

\subsection{Comparators} The whole-budget solo gives all $T$ bits to the sender chosen as stronger on selection data, which sends its class label through a code and a fitted receiver. Where stated, the solo is the classifier's original prediction (its native decision) or a code of its 32 posterior clusters. The fusion reference combines the senders' full class-probability vectors by the product rule~\citep{Kittler1998}, and a full-input model is fitted to all views at once. A sender counts as capable when its accuracy on selection data, or on the development objects a study declares for this purpose, reaches 80 percent, the competence threshold.

\subsection{Timing of Predictions and Analyses} These analyses were not formally preregistered. Predictions for the body-sensor, end-to-end, Dry Bean and complementary-views analyses were specified before their test predictions were evaluated. For the bird and research-paper tasks, they were specified before feature extraction. Predictions for the six-view analysis were specified before its four additional views were examined, although the same test objects had already been evaluated in the two-view analysis. The language-model analysis fixed its deployment rule before test labels were read, and the quantized 7B analysis fixed its competence requirements before evaluation. The predictions and their outcomes are given in the SI Appendix (sections~S16, S17, S18, S19 and~S20).

\subsection{Statistics and Reproducibility}\label{sec:repro} Comparisons within a dataset are paired on one shared test set. Means, minima, maxima and counts of units with a positive difference describe that finite test set, and spreads over units, persons and training splits are descriptive. The MNIST interval resamples test images jointly across the frozen models with 100,000 draws. The Dry Bean and 7B intervals use 2,000 paired-object bootstrap draws with all fits frozen. The language-model study's deployment rule uses a simultaneous Hoeffding margin over its declared comparisons~\citep{Hoeffding1963}.

We also checked computational reproducibility by repeating receiver fitting and selection from saved model outputs, using the same data. Details of these checks and the numerical verification of the quantized 7B implementation are given in SI Appendix, sections~S14 and~S20.

\section*{Data and Code Availability}
All datasets and pretrained models are publicly available. The datasets comprise optical digits~\citep{AlpaydinKaynak1998}, pen digits~\citep{AlpaydinAlimoglu1996}, Multiple Features~\citep{Duin1998}, MNIST~\citep{LeCun1998}, Dry Bean~\citep{DryBean2020}, Daily and Sports Activities~\citep{UCIdsa}, PAMAP2~\citep{Reiss2012,UCIpamap}, Breast Cancer Wisconsin (Diagnostic)~\citep{UCIwdbc}, hierarchical DBpedia~\citep{DBpediaHierarchy}, CUB-200-2011 photographs and attributes~\citep{Wah2011CUB} with Reed et al.'s captions~\citep{Reed2016}, and Web of Science~\citep{WOSdata}.

Pretrained models were obtained through the Hugging Face Hub: SmolLM2 at 135M and 1.7B parameters~\citep{SmolLM2025}, Qwen2.5 at 0.5B, 1.5B and 7B parameters~\citep{Qwen2025}, and DINOv2 small and base~\citep{Oquab2024DINOv2}.

Analysis and figure-generation code are available from the author as a code-only reproducibility package, including source URLs, exact model revisions, computational environments and instructions for obtaining the data. No datasets or model weights are redistributed.

\section*{Acknowledgement}
This work was supported by the Deutsche Forschungsgemeinschaft (German Research Foundation) under Germany's Excellence Strategy---EXC 2117--422037984.

AI assistance from Anthropic Claude Code (Opus 5, Opus 5.5, Fable 5 and Fable 5.1) and OpenAI ChatGPT/Codex (GPT-5.6 Sol and GPT-6 Astra) was used for readability suggestions and language edits, programming, symbolic and algebraic checks, plotting codes, literature and dataset searches, manuscript review, and proofreading the manuscript. The author verified all AI-assisted input and takes full responsibility for the content, analyses and conclusions.

\section*{Author Contributions}
M.S. designed and performed the research and wrote the paper.

\section*{Competing Interests}
The author declares no competing interests.

\clearpage

\clearpage

\clearpage

\clearpage

\clearpage
\clearpage
\end{document}